\documentclass{article}
\usepackage[english]{babel}
\usepackage[a4paper,top=2cm,bottom=2cm,left=3cm,right=3cm,marginparwidth=1.75cm]{geometry}
\usepackage{amsmath}
\usepackage{amsfonts}
\usepackage{graphicx}
\usepackage{tcolorbox} 
\usepackage[colorlinks=true, allcolors=blue]{hyperref}
\usepackage{url}
\usepackage{natbib}
\title{Conditional updates of neural network weights for increased out of training performance}
\author{Jan Saynisch-Wagner$^{1,\dagger}$ and Saran Rajendran Sari$^1$}

\date{%
    $^1$GFZ Helmholtz Centre for Geosciences, Potsdam, Germany\\%
    $^\dagger$\texttt{e-mail: saynisch@gfz.de}%
}

\begin{document}
\maketitle

\begin{abstract}
This study proposes a method to enhance neural network performance when training  data and application data are not very similar, e.g., out of distribution problems, as well as pattern and regime shifts. The method consists of three main steps: 1) Retrain the neural network towards reasonable subsets of the training data set and note down the resulting weight anomalies. 2) Choose reasonable predictors and derive a regression between the predictors and the weight anomalies. 3) Extrapolate the weights, and thereby the neural network, to the application data. We show and discuss this method in three use cases from the climate sciences, which include successful temporal, spatial and cross-domain extrapolations of neural networks.
\end{abstract}

\section{Introduction}
In physics, especially in geosciences and climate sciences, the poor performance of neural networks (NN) when applied outside their training distribution or their trained dynamics poses a very strong limitation to their general applicability \citep{Irrgang.etal.2021,Landsberg.Barnes.2025}. 

In these fields, physical relations such as laws, dependencies or sensitivities are commonly derived (or learned) under well observed conditions and are then applied to less observed conditions to gain knowledge about the latter. For example, results from lab or numerical model experiments are regularly applied to real world problems or observations  \citep[e.g.,][]{Mehta.etal.2025}; knowledge from our Earth and our Solar System are transferred to other planets and other star systems \citep[e.g.,][]{Sachl.etal.2026}; learned relations that are derived today are transferred to the distant past or to the future \citep[e.g.,][]{CMIP6,Wang.etal.2024,Koutsodendris.etal.2014}. Usually, well-documented problems are transferred, e.g., across country borders \citep{Sebastianelli.2024,Kotz.etal.2021}, across spatial scales \citep{Ortega-Cisneros.etal.2024}, from the well observed upper ocean to the deep ocean \citep{Llovel.etal.2014,Lee.Gentemann.2018}, or from the lower latitudes to the poles \citep{Irrgang.etal.2020}; or a mixture of it all \citep{Jung.etal.2024,vdDeure.etal.2025}.

From a machine-learning perspective, this means that available training data are typically limited in their spatial and temporal extent, as observations or model simulations only cover specific time periods or regions.
For example, Earth observing satellites often leave the higher latitudes unobserved depending on their inclination; oceanographic ARGO floats measure only the upper 2000 m of the marine water column; paleo records such as ice cores may exist for the past but only in regions where there is still ice today. In general, the majority of measurements cover only the last few decades, no measurements exist for the future and comparatively few measurements exist for the distant past. Furthermore, the environmental conditions that exist on Earth are very heterogeneous and most processes, for instance the climate, are non-static in nature. The combination of spatiotemporally-localized data distributions in combination with differing conditions between possible inference regions/times/environments and the application regions/times/environments makes it very difficult for NN to operate. 

In some cases, individual standardization or rescaling of the variables from the application data set is a valid option to mitigate the out of distribution (OOD) problem. However, in many real use cases, the standardization or rescaling removes essential information from the application data. This is true for nonlinear problems where standardizing or rescaling the input will lead to wrong outputs as by definition of non-linearity $f(a+b) \neq f(a) + f(b)$ and $f(n*a) \neq n*f(a)$. Especially in the climate sciences the conditions for contemporary AI are challenging and raise the demand for new approaches \citep[e.g.,][]{Eyring.etal.2024,Fang.etal.2025}. For example, removing offsets and trends in variables such as sea level and temperature would discard information essential for predicting climate change, climate impacts, and tipping. Likewise, the amplitudes of extreme events should not be rescaled to fit that of the training data. In addition, standardization and rescaling requires a certain amount of data points, which are not always available. Often, the learned relations have to be applied to one single (possibly extreme) event, to a few ice cores, to one other planet, moon or star system and so on, i.e., they pose so called single-shot or few-shot problems in AI terminology. Consequently, in such cases a standardization or rescaling is not possible.   Finally, even without any distribution shift or with normalization, the NN can still be confronted with unseen combinations of input values which can pose a serious OOD problem. \\

Here we propose a method to reduce this problem by exploiting the NN's own sensitivity towards samples of the training data to predict NN weights and biases which are intended to perform better in the OOD application realm (cf. Sec.~\ref{sec:methods}, Method). 

We believe that many fields of physics and other fields of science as medicine, biology, economic sciences or social sciences can benefit from the proposed method. Nonetheless, we demonstrate and evaluate the proposed methodology in three classes of experiments that are mainly motivated by climate sciences and Earth sciences (cf. Sec.~\ref{sec:experiments}, Experiment Design and Data). One experiment, where NN weights are predicted through time, one experiment where NN weights are predicted through space, and one experiment where NN weights are predicted across domain-boundaries. Within similar training and application data sets and domains, the chosen experiments do not represent any challenges to NN training (nor to basic mathematics and physics). The challenges we demonstrate and mitigate here arise only by training the NN in times and places that have very different conditions than the times and places the model has to operate on during application. Consequently, these experiments should be understood entirely as symbols or place-holders for more sophisticated AI applications. 

\section{Experiment Design and Data}
\label{sec:experiments}
\subsection{Experiment 1: Tipping of the Atlantic Meridional Overturning Circulation}
\label{sec:exp1}
We base experiment 1 on \cite{vanWesten.etal.2024}, a paper that shows that critical weakening of the Atlantic Meridional Overturning Circulation's (AMOC) strength can occur in complex climate models, too. This so-called AMOC tipping is triggered by forcing the northern Atlantic Ocean of the respective climate model with increased freshwater flux (e.g., related to the rapid melting of the Greenland Ice Sheet). These so-called hosing experiments are a common tool in climate sciences \citep[e.g.,][]{Rahmstorf.1996,Saynisch.etal.2016}. The paper's supplementary material and the data we use here is located in the following repository: \url{https://github.com/RenevanWesten/SA-AMOC-Collapse/tree/SA-AMOC-Collapse_v1.0} 

As input to the NN, we use 2d velocity fields representing vertical slices through the Atlantic Ocean, i.e., the zonal velocity maps (VVEL). These model based horizontal velocity maps are taken always at the same location and are given as annual values. The VVEL data can be downloaded from the above given repository from the folder:
\path{/Data/CESM/Data/AMOC_section_26N/}. The files cover each 50 years of data, the first one is called \path{CESM_year_0001-0050.nc} and so forth.

The target output of the NN is the strength of the AMOC given in Sverdrup (1Sv = $1^6 m^3/s$), i.e., a single value per input velocity map. The AMOC strength is defined as the total meridional volume transport at 26$^\circ$N over the upper 1000 m (cf. eq. 3 in \cite{vanWesten.etal.2024}). The respective time series is given in \path{/Data/CESM/Ocean/AMOC_transport_depth_0-1000m.nc}

The challenge in experiment 1 is that the training bases only on data before the AMOC tipping point, occurring around year 1800 of the model data, cf. Fig.~\ref{fig:AMOC}. However, the NN has to calculate the correct AMOC strength also during tipping and even after tipping, a serious OOD problem.

\subsection{Experiment 2: Estimation of Sea Water Density}
\label{sec:exp2}
Experiment 2 bases on the highly nonlinear Equation Of State for sea water (EOS).
For every desired oceanic location, the NN input is salinity (S), temperature (T),    
pressure (P), latitude ($\phi$), longitude ($\lambda$) and depth (z) and the target output is the local in-situ density ($\rho$) at this particular location. Consequently, the NN task is to convert 6 numbers into one.
For data we use the gridded observations of the World Ocean Atlas \citep{WOA23_T,WOA23_S} which are available under \url{https://www.ncei.noaa.gov/access/world-ocean-atlas-2023/}. We use the 5$^\circ$ decadal averaged annual fields of S and T for every available latitude (36 levels), longitude (72 levels), and depth (102 levels) of the provided 3d ocean grid. These values represent long term averages and no temporal dimension is present in this experiment. We then calculate P and $\rho$ by using  the routines \texttt{swPressure(z,$\phi$)} and \texttt{swRho(S,T,P)} from the \texttt{oce} R-package \citep{OCE} which follow the Thermodynamic Equation Of Seawater - 2010 (TEOS-10) standards \citep{TEOS10}.

Experiment 2 is motivated by the diminishing of oceanic monitoring capabilities with depth. Let's assume that we have found an interesting relation between our targets and the ARGO float data. Unfortunately, the ARGO floats cover only the upper 2000 m of the water column. However, we want to exploit our findings also below 2000 m. In a practical application case this relation could estimate krill, CO$_2$ or nutrients. Here, we  restrict ourselves to the estimation of sea water density. 

In contrast to experiment 1 where knowledge has to be transferred over time (actually, also over to different post-tipping dynamics), the challenge of experiment 2 is to transfer learned knowledge spatially. The training of the NN is restricted to a spatial subset of the world oceans. However, the trained NN has to redo the calculation for the entire ocean where OOD input and output values can occur. A further difficulty is that the EOS follows a complicated, highly-nonlinear empirical polynomial \citep[cf.][]{Roquet.etal.2015}.

\subsection{Experiment 3: Uncertainty Estimation of Global Wind Velocity Reanalyses}
\label{sec:exp3}
The third experiment focuses on cross-domain extrapolation of learned uncertainties in global wind velocity reanalyses. The setup follows the concept proposed by \citet{Irrgang.etal.2020}, who demonstrated that spatiotemporal uncertainty patterns learned at a single oceanic location can be generalized to the global ocean domain. Here, we extend this approach by transferring the learned uncertainty information from an oceanic reference point to continental regions. This setup represents an OOD scenario, since wind dynamics over land are governed by different boundary-layer processes, surface roughness, and convective regimes compared to those over the ocean.

In contrast to \citet{Irrgang.etal.2020}, who used three atmospheric reanalysis products as model input, we restrict the input data to two widely used reanalyses: ERA5 \citep{ERA5} from the European Centre for Medium-Range Weather Forecasts (ECMWF) and CFSv2 \citep{Saha.etal.2014} from the National Centers for Environmental Prediction (NCEP). Data set preparation, NN model architecture, and NN model training follow the same strategy as in the reference study to ensure comparability.
This configuration tests whether the learned sensitivities in wind velocity uncertainty over marine environments can meaningfully generalize to terrestrial conditions, thereby illustrating the broader potential of the spatial transferability of the proposed method.

\section{Method}
\label{sec:methods}
The proposed weight-prediction method for NN follows the pseudo-code given in Tab.~1. At its core, the approach consists of 3 parts. After the training of a suitable NN to the problem at hand with a training data set (Tab.~1, step 1), one has to do the following. A: For each weight and bias of this hereafter termed "parent model", collect the weight-deviations arising from retraining the parent model with single data points of the training data set (Tab.~1, step 2). B: For each weight and bias of this parent model, establish a dependency between said deviations and suitable predictors, e.g., a linear regression (Tab.~1, step 3). C: For each weight and bias of this parent model, extrapolate the established dependency towards target data points and thus generate new target data tailored models, the "child models" (Tab.~1, step 4). From now on, we summarize a group of child models originating from the same trained parent model under the singular "child model". All models in one child model differ only due to the fact that their weights are predicted onto different target data points (cf. $x_t$ and \textit{ChildModel$_t$} in Tab.~1, step 4), until this step they base on exactly the same data. Consequently, if we use the plural from now on, we refer to different child model groups, e.g., each originating from a different parent model.

Note, in  Tab.~1 we show a total-valued form of the method. Naturally, handling only the weight anomalies or increments $\Delta$\textit{W$_{ik}$} is equally possible. 

\begin{tcolorbox}[
colframe=blue!25,
colback=blue!10,
coltitle=blue!20!black,  
fonttitle=\bfseries,
adjusted title=Table 1: Weight-Prediction Approach (pseudo-code)]
\begin{enumerate}
\item \textbf{Train}: 
\begin{itemize} 
\item \underline{Train} suitable NN on training data set $\mathbb{I}$, consisting of $I$ data points $x_i$ 
\item \underline{Store} \textit{ParentModel$_0$}
\end{itemize}
\item \textbf{Forgetful online learning}: 
\begin{itemize} 
\item \underline{Loop} over a reasonable subset of training data points $x_i$, e.g., for all $x_i \in \mathbb{I}$:
 \begin{enumerate}
     \item \underline{Finetune} \textit{ParentModel$_0$} on single training data point $x_i$ regularized towards $\mathbb{I}$. For example, by compiling the $x_i$-focused training data set $\mathbb{I}_i$ consisting of n copies of $x_i$ and n random other training data points  $x_{j\neq i} \in \mathbb{I}$.\\
     Result: \textit{ParentModel$_i$}
     \item \underline{Store} \textit{ParentModel$_i$}'s  $K$ weights (including biases) of all layers, nodes and connections:  \textit{W$_{ik}$}
     \item \underline{Reload} \textit{ParentModel$_0$}
     \item \underline{Repeat} from (a)
 \end{enumerate}
\end{itemize} 
\item \textbf{Weight-Regression}: 
\begin{itemize} 
  \item \underline{Choose} suitable regression algorithm, e.g., linear regression \texttt{lm()}
 \item \underline{Choose} (and prepare) a respective set of predictors/regressors $\mathbb{R}$, which correspond to the training data points $x_i$ used in step 2 above and the target data point(s) $x_t$ used in step 4 below: $R_i$ \& $R_t \in \mathbb{R}$, e.g., time steps $t_i$ \& $t_t$
\item \underline{Loop} over a suitable subset of the \textit{ParentModel$_0$}'s $K$ weights and biases, e.g., for all $k \in \mathbb{K}$:
 \begin{enumerate}
     \item \underline{Fit} chosen regression model, e.g., \textit{RegressionModel$_{k}$} = \texttt{lm(}\textit{W$_{ik}$}, R$_{i}$\texttt{)}
     \item \underline{Store} \textit{RegressionModel$_{k}$}
     \item \underline{Repeat} from (a)
 \end{enumerate}
 \end{itemize} 
 \item \textbf{Weight-Prediction}: 
\begin{itemize} 
\item \underline{Loop} over a reasonable subset of target data points $x_t$, e.g., for all $x_t \in \mathbb{T}$:
  \begin{enumerate}
         \item \underline{Predict} all required weights for target data point $x_t$:  \textit{W$_{tk}$} = \textit{RegressionModel$_{k}$}(R$_t$)
         \item \underline{Store} \textit{ChildModel$_t$}
         \item \underline{Repeat} from (a)
 \end{enumerate}
 \end{itemize} 
 \item \textbf{Actual Application}: \begin{itemize} \item \underline{Apply} child model to target data point(s): \textit{ChildModel$_t (x_t)$}
 \end{itemize} 
\end{enumerate}
\end{tcolorbox}

The exact execution of each of the mentioned steps can vary. However, some main choices have to be made during the approach in any case.
First, choices for suitable predictors for the weight-regression (Tab.~1, step 3) and the subsequent weight-prediction (step 4), have to be made. It is out of scope of this paper to discuss these choices here in detail. Ideally the choices should be heavily influenced by prior knowledge of the problem at hand and how it may behave outside the training data range. In principle, the predictors for the weights can be anything, e.g., i) time, ii) (subsets of) the original NN input, iii) features derived from the original input, or iv) additional information not used by the NN model. A careful selection and pruning of the predictors is advised.
In our examples we demonstrate the use of ii) in experiment 2, and iii) in experiment 1 and 3. Note that in experiments 2, the inputs of depth, latitude and longitude are not necessary for the NN to calculate the correct sea water density as long as temperature, salinity, and pressure are given. However, they are promising weight-predictors. As we did not want to limit the NN by withholding information, we unified the inputs among NN training and weight-regression in experiment 2.

Second, a choice of weight-prediction method has to be made. Of these, many suitable exist, e.g., auto-regressive integrated moving average (ARIMA), XGboost, linear models (LM), generalized additive model \citep[GAM,][]{Hastie.Tibshirani.1986}, random forest \citep{Ho.1995}, NN itself \citep[cf.][]{Chauhan.etal.2024} and many more. Again, it is not part of this paper to review regressions methods and their rightful application. In experiments 1 and 2, we choose LM for their ease of use and very fast computation. The latter is especially needed as we do regression and prediction for every weight and bias of every single node of the NN. However, if the NN is very large, low-rank \citep[LoRA,][]{Hu.etal.2022} or similar decompositions should be considered. We would not generally recommend to use other purely data driven methods like RF and NN as they can suffer from poor OOD performance, too. Still in some cases we found improvements in our experiments by these methods. For instance, in experiment 3, we demonstrate the use of a NN for the weight-regression and weight-prediction tasks.

\section{Results and Discussion}
\subsection{Experiment 1: Tipping of the Atlantic Meridional Overturning Circulation}
To predict AMOC strength from velocity maps, we use a standard shallow convolutional NN (CNN), with 2 convolutional/max pooling layers followed by two dense layers. The last layer has linear activation while all other layers have the Exponential Linear Unit (ELU) activation. For the  online learning (Tab.~1, step 2a), n is set to 200.

For the regression of the NN weights, the predictors are based on the decomposition into empirical orthogonal functions (EOF) of the training data itself. The principal components of the 4 leading EOF, i.e., 4 numbers per time step, are used for the regression of the NN weights. Outside the training data, exactly the same EOF-basis as during training is used to decompose the respective velocity maps. Another possible set of predictors could be mean, maximum, minimum and standard deviation of each individual velocity map, i.e., again 4 numbers per time step, or even time itself if a suitable low dimensional bifurcation model were to be employed for the NN weight-prediction \citep[cf.][]{Rahmstorf.1996}. Note however, that the sensitivity of the weights is not dictated by the domain dynamics alone but also depends on the involved activation functions (and not only the activation function from the node under consideration). 

As there is a clear order of the data in experiment 1, i.e., the temporal order, we skip the "reload ParentModel$_0$" step of the approach (cf. Tab.~1, step 2c). This way, the online learning is not forgetful and the weights will evolve naturally along time. This improves the results in this experiment as it reduces noise in the weight sensitivities (\textit{W$_{ik}$}) generated during step 2.
After establishing weight-predictor relations for every weight and bias of the CNN, the child model was predicted with sub-models (cf. \textit{ChildModel$_t$}) for every single time step of the problem by using the same weight-predictor relations. As a consequence, each of these sub models was then applied only to one dedicated data point.

\begin{figure}
    \centering
\includegraphics[scale=0.6,trim=0cm 0.5cm 1cm 1.5cm,clip=true]{
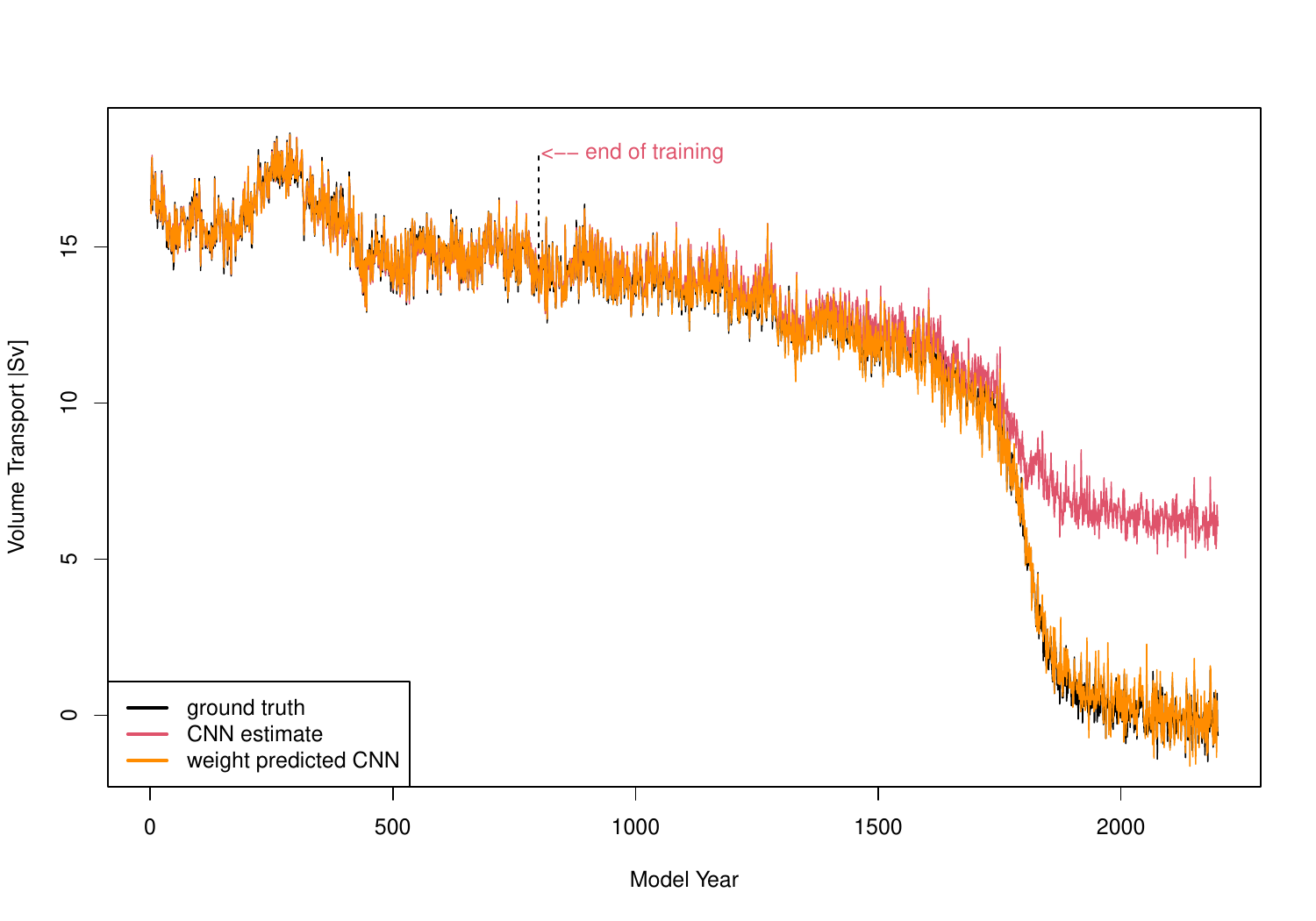}
   \caption{AMOC strength and an example of good performance of the weight-prediction method. Black Line: AMOC overturning decay and tipping, i.e., the target ground truth values. Red Line: Output of parent model trained until the dashed end of training (EOT) line. Orange Line: Output of weight-predicted child model. Forgetful online learning and subsequent fitting of EOF-to-weight relations use only data before dashed EOT line.}
    \label{fig:AMOC}
\end{figure}

Figure~\ref{fig:AMOC} shows the time series of the target value (black line), the output of the parent model (red line), and the output of the same model after the application of the weight-prediction approach (orange line), i.e., the child model. The training data starts at year 1 and the end of the training data (EOT) is shown by a vertical dashed line, here at 800 years.   
After the tipping of the AMOC sets in at around 1800 years, the parent model has clearly  problems reproducing the black line. Its output is strongly biased in the direction of the output it had learned during training. The child model surpasses the parent model in its performance during and after tipping.

Please note, that since ParentModel$_0$ was not reset during step 2 of Tab.~1, at the end of step 2 we get a NN model that has seen the training data more often than ParentModel$_0$. Still the performance of this model (not shown) is the same as that of ParentModel$_0$ and the improvements plotted in Fig.~\ref{fig:AMOC} must be attributed entirely to steps 3, 4, and 5 of our approach.

\begin{figure}
    \centering
\includegraphics[scale=0.46,trim=0.1cm 0.5cm 1cm 1.5cm,clip=true]{
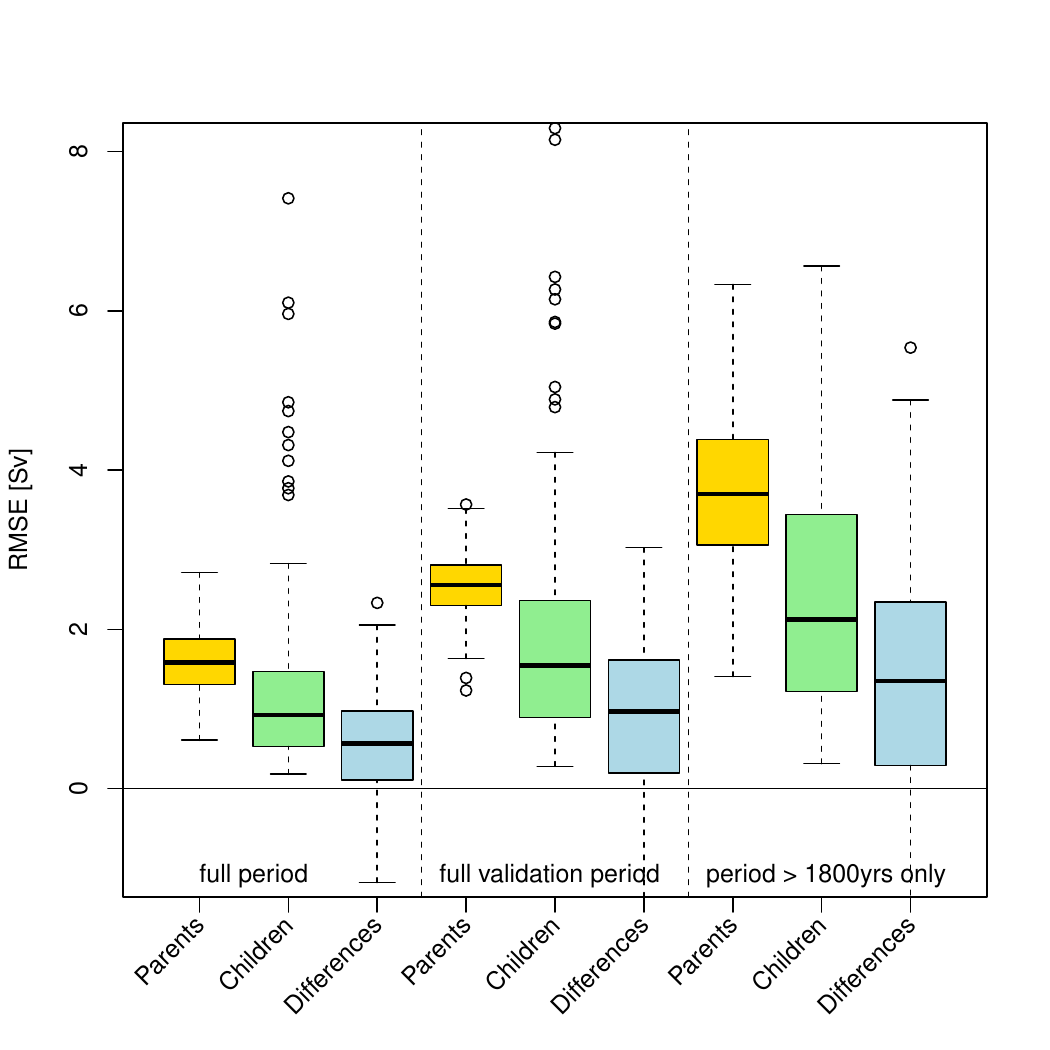}%
\includegraphics[scale=0.46,trim=1.5cm 0.5cm 1cm 1.5cm,clip=true]{
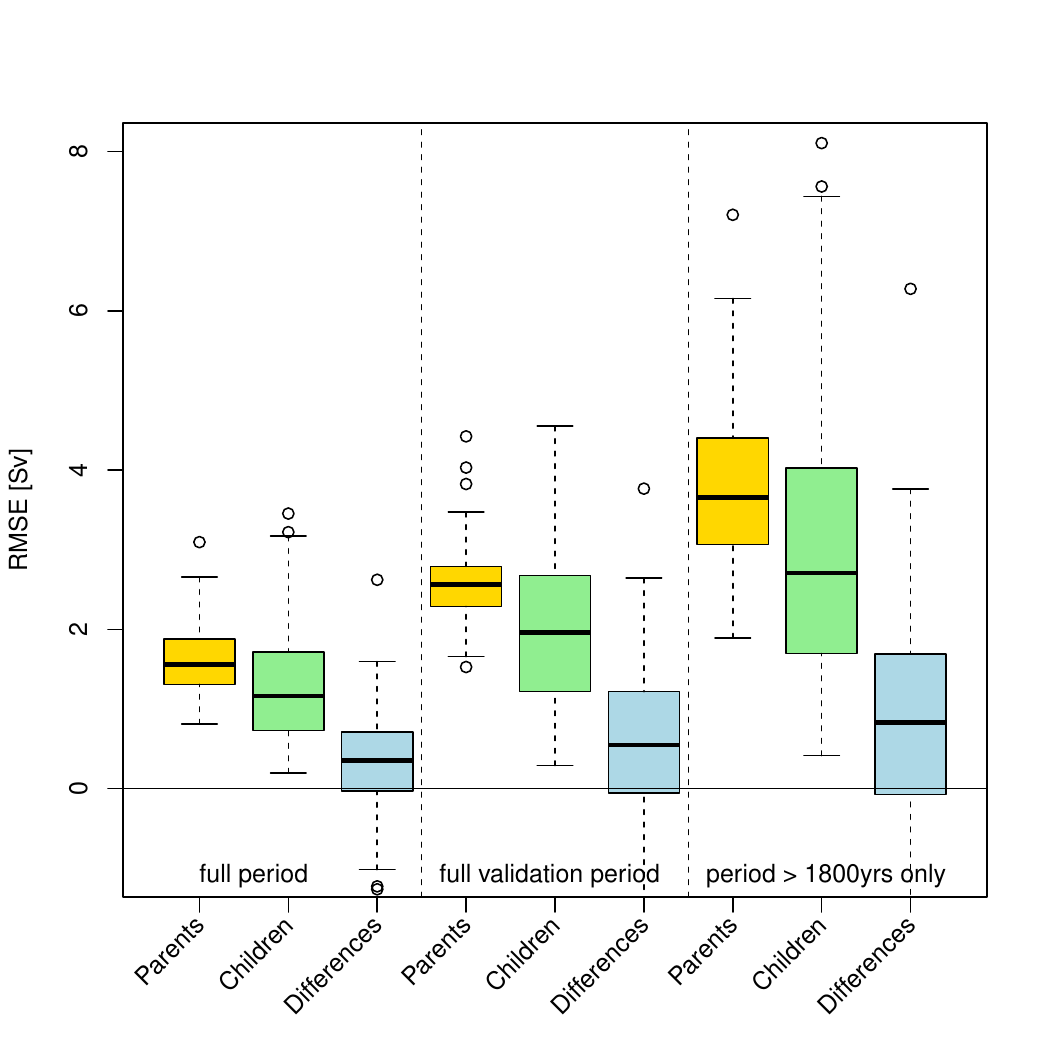}%
    \caption{Impact of the choice of the weight-prediction method on the predicted model's performance. Polynomials of order 1 (left panel) and 2 (right panel) where fitted to the leading 4 principal components of the input variable (i.e., the meridional velocity maps). Yellow Boxes: RMSE of unperturbed models. Green Boxes: RMSE of weight-predicted models. Blue Boxes: Pairwise RMSE differences (i.e., unperturbed "parent" model minus corresponding weight-predicted "child" model). Positive values represent improvements due to the presented weight-prediction approach. Note, "full period" and "period \textgreater 1800" is the same for every simulation of the ensemble. The term "full validation period" refers to all data point after each simulation's individual EOT (which is randomly chosen for every simulation before training). Each box bases on 300 parents/children/pairs.}
    \label{fig:poly12}
\end{figure}

Since in general the training of NN can have largely varying results, we repeat the experiment several hundred times with varying EOT times, while the beginning of the training window remains fixed at year 1. As a consequence, the amount of data each NN was trained on varies as well as how close the EOT comes to the actual tipping event. Furthermore, two weight-regression methods are tested. The results are aggregated in Fig.~\ref{fig:poly12}.

The figure shows root mean squared errors (RMSE) of parent models (yellow boxes), child models (green boxes), and their pairwise differences (blue boxes). For a successful  weight-prediction, 
the child models should show lower RMSE than the parent models and the pairwise differences should show as positive values as possible. The variations summarized by each box bases on the rather random generation of parent models in combination with random selection of an EOT between years 800 and 1800. In each panel, three horizontal subdivisions are shown where the RMSE are calculated for 3 different time windows: the full 2200yrs of data including the training data (left), the time beyond each NN's individual EOT, i.e., everything except the training data (middle), and a fixed time window covering years 1800-2200, i.e., the tipping event and after (right).  

The two panels of Fig.~\ref{fig:poly12} are both based on a linear model for the weight-regression. The standard R function \texttt{lm()} was used here. The difference between the two panels is the order of polynomials built from the predictors before weight-fitting. While the left panel uses polynomials of order 1, the right panel uses polynomials of order up to 2.
As mentioned, the weight-regression bases on EOF-decomposition of the 2d meridional velocity fields that form the input for the NN itself. The predictors are always the 4 principal components associated with the 4 leading EOF as derived in the training time window. 

One can see, that in all cases the (green) child model boxes show lower RMSE than the (yellow) parent model boxes, leading to positive pairwise RMSE differences depicted as blue boxes. In addition, the performance gain of the child models compared to their parents increases when the models face conditions that are increasingly different from their training data. The average RMSE of both, parent and child, boxes become larger from left to right in each of the two panels. From left to right, the RMSE in the subdivisions increasingly focus on the OOD performance of the NN. However, the RMSE increase stronger in the parent models than in the child models resulting in larger pairwise improvements when the tipping is approached and surpassed. 

Since the regressions of  polynomials of order 1 and order 2 give comparable results with no clear preference so far, this could be interpreted as a sign of robustness of the approach towards the regression method. However, polynomials of even higher order (as far as tested) gave worse results (not shown). Here, the higher order terms get random/noise-based slopes/gradients assigned to them during the regression (Tab.~1, step 3) as the resulting terms have only negligible influence within the training data (where the regression is conducted). Nonetheless, these terms can still lead to significant yet wrong influences during and after tipping in step 4 of Tab.~1. More research on how to stabilize the child generation is needed. A signal-to-noise ration guided pruning of insignificant slopes returned by the regression step could help with this problem.

In general, as can be seen in Fig.~\ref{fig:poly12} the approach's success can vary considerably, resulting even in worse results of some child models.  The same mechanism described above for higher polynomials probably is the reason for the poor results in some of the polynomials of order 1 and order 2 cases, too. On average, the results are positive as the ensemble mean of the child models shows a clear improvement over the ensemble mean of the parent models. More research is needed to stabilize the results, either by identifying parent models unfit for the approach, or if possible, by subjecting child models to forms of validation before their application. Until then, an ensemble approach should be considered.

In the following, we will discuss some noteworthy examples from the ensemble simulations aggregated in Fig.~\ref{fig:poly12}.  Apart from the many good results that look more or less similar to Fig.~\ref{fig:AMOC}, it can happen that the regression shows too little sensitivity to the predictors and the parent and child model outputs are very similar.

How does it come to this poor performance is rather unclear at the moment and needs more investigation. The most likely explanation is that the respective parent model sits either in a deep narrow minimum or a very flat part of the loss hyperplane and due to this, the proposed forgetful online learning resulted only in minimal variations of the weights ($\Delta$\textit{W$_{ik}$}). As the particular parent model is not sensitive to the online training step (Tab.~1, step 2), the weight-regression and the subsequent weight-prediction inherit this limitation. Still, the small induced weight variations are not pure noise and the weight-prediction generates a better performing child model, even if it is very similar to its parent (not shown). 

\begin{figure}
    \centering
\includegraphics[scale=0.6,trim=0cm 0.5cm 1cm 1.5cm,clip=true]{
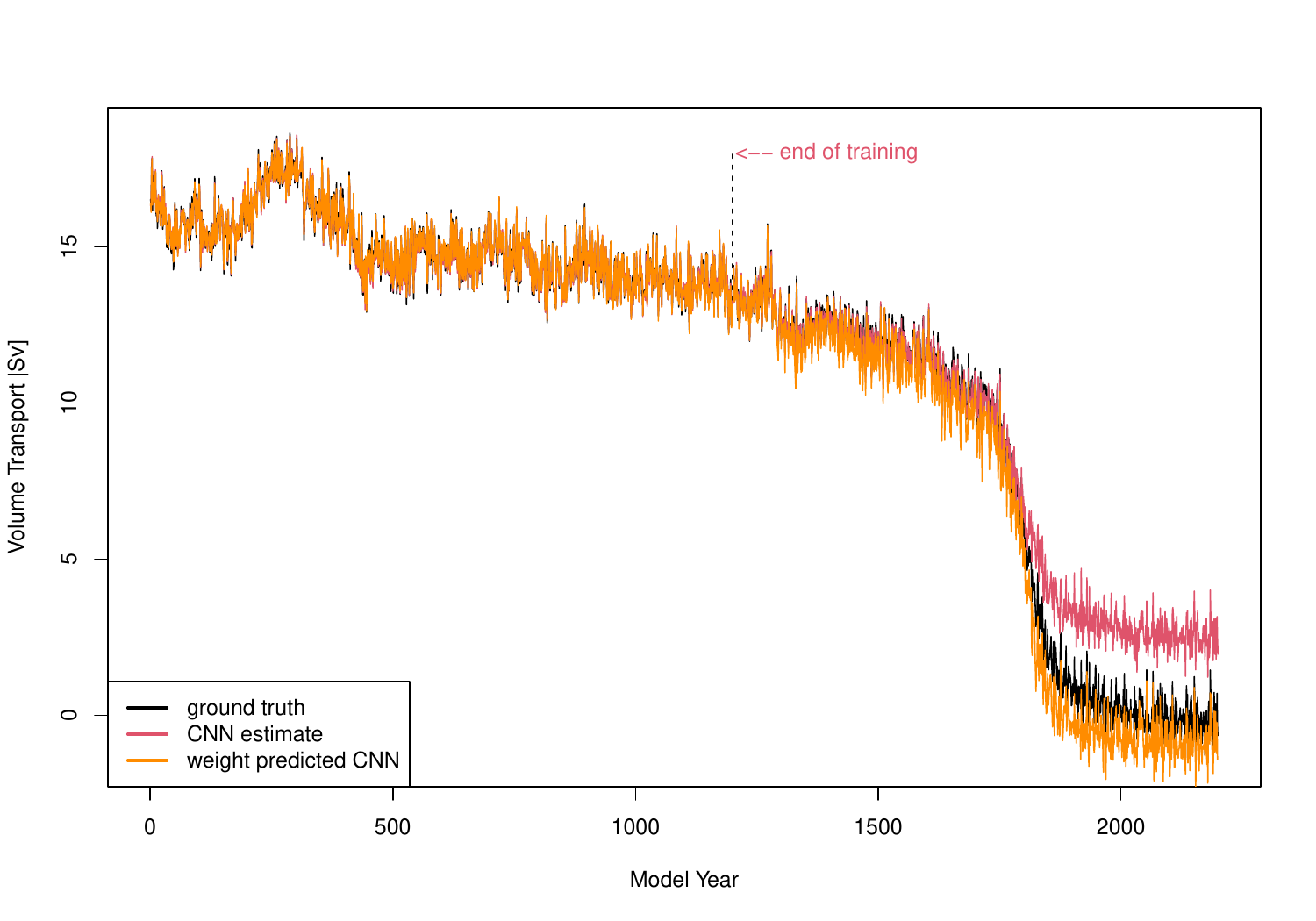}
   \caption{As Fig.~\ref{fig:AMOC}, but with a sub-optimal example of child model performance. Still, the predicted child model (orange) outperform the parent model (red) beyond the EOT threshold in the RMSE sense.}
    \label{fig:fail}
\end{figure}

Furthermore, in some rare cases deterioration can happen when a child model is applied to data similar to the training conditions as shown in Fig.~\ref{fig:fail}. The child model (orange line) bends too early towards the post-tipping output and then overshoots. As a consequence, in the 500 year period following the EOT (at year 1200) the parent model (red line) slightly outperforms the child model (orange line). When calculating the RMSE for the entire post-EOT time, the child model still outperforms its parent model.

 \begin{figure}
     \centering
 \includegraphics[scale=0.46,trim=0.1cm 0.5cm 1cm 1.5cm,clip=true]{
 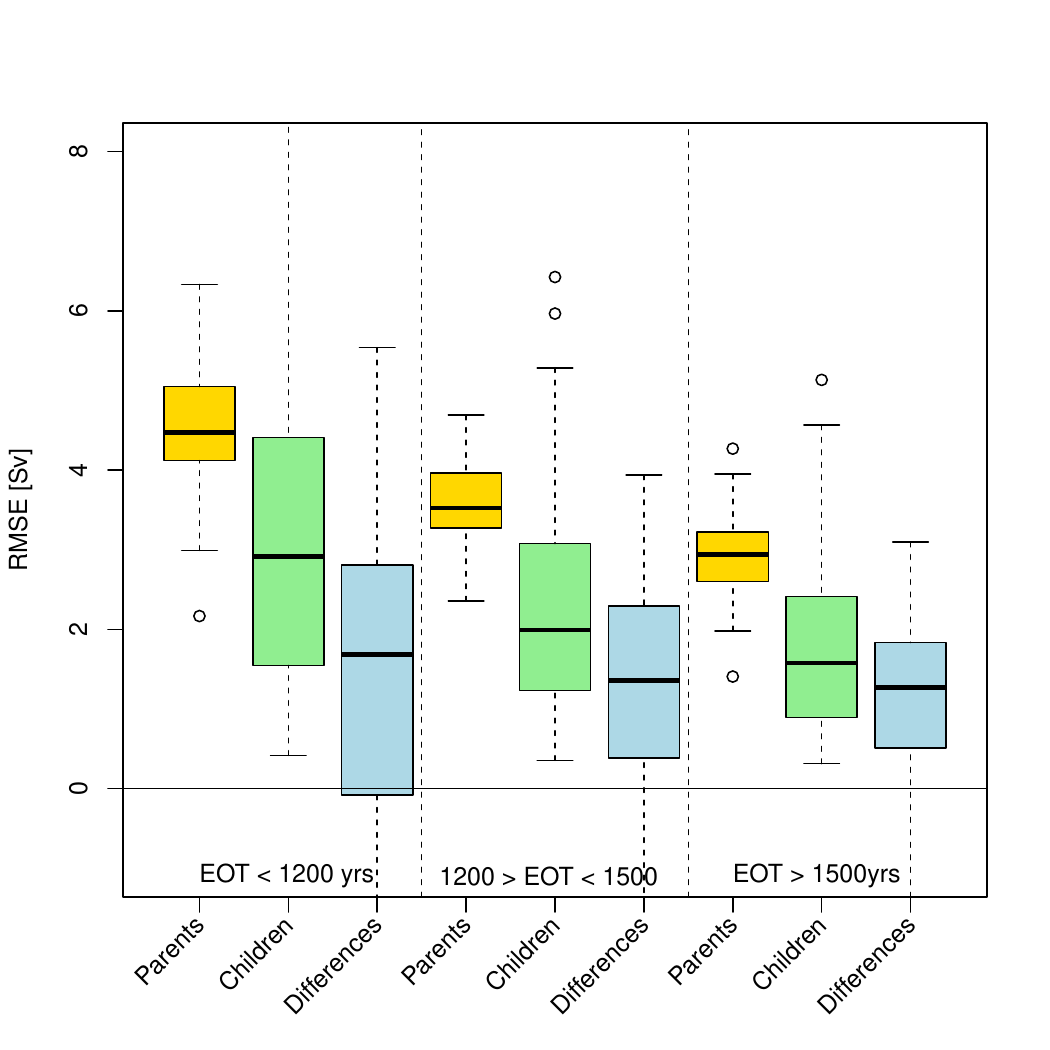}%
 \includegraphics[scale=0.46,trim=1.5cm 0.5cm 1cm 1.5cm,clip=true]{ 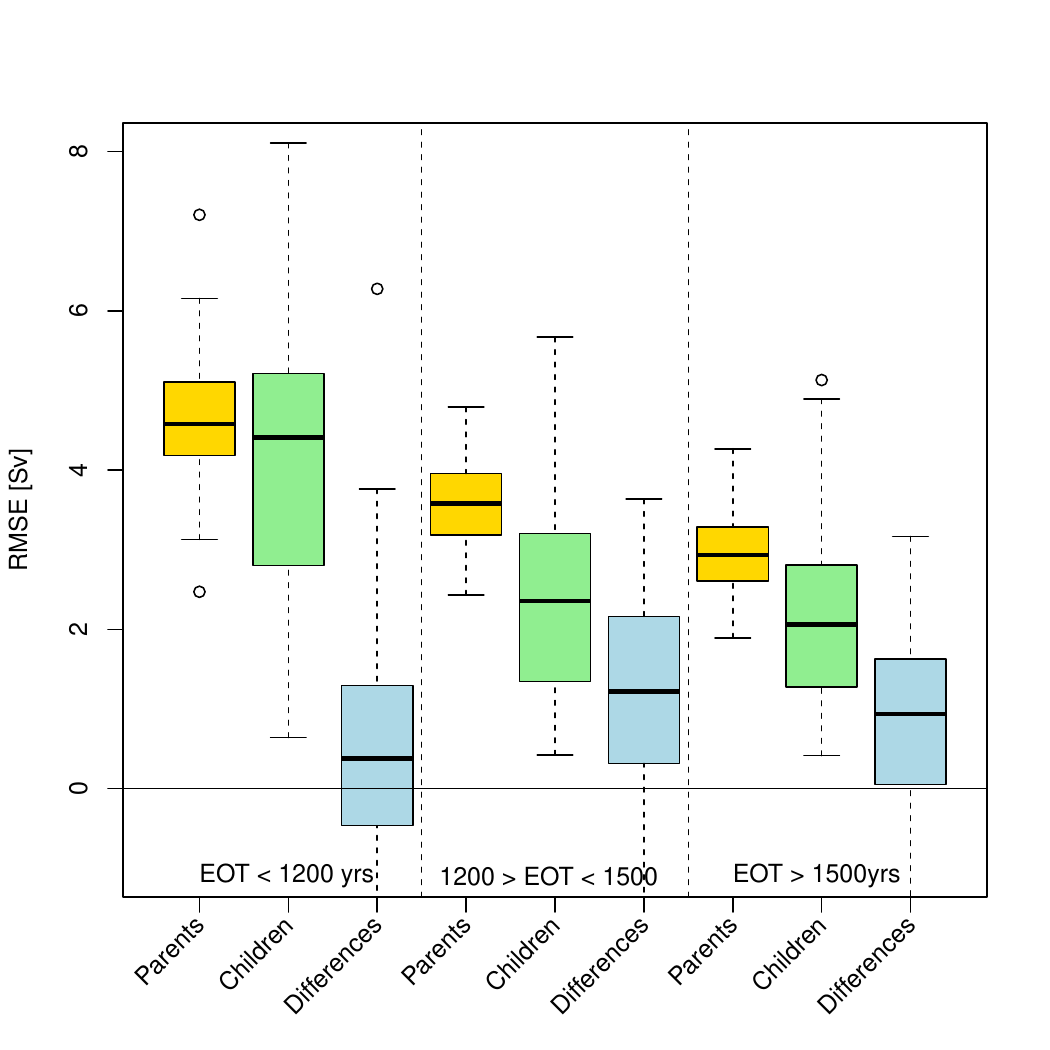}%
     \caption{Impact of EOT on the RMSE of a fixed target range (1800yrs,2000yrs]. Polynomials of order 1 (left panel) and 2 (right panel) where fitted to the leading 4 principal components of the input variable (i.e., the meridional velocity maps). Yellow Boxes: RMSE of unperturbed models. Green Boxes: RMSE of  weight-predicted models. Blue Boxes: Pairwise RMSE differences (i.e., unperturbed "parent" model minus  corresponding weight-predicted "child" model). Positive values represent improvements due to the presented  weight-prediction approach. Each box bases on 100 parents/children/pairs.}
     \label{fig:EOT_split}
 \end{figure}

Figure~\ref{fig:EOT_split} bases on the same data as Fig.~\ref{fig:poly12}, but here the RMSE are sorted by EOT. In addition, the figure shows only the RMSE for the tipping and past tipping period. While in the view of Fig.~\ref{fig:poly12}, the polynomial of order 1 and order 2 based regressions did look very similar, here in Fig.~\ref{fig:EOT_split}, some large differences become evident. If trained only very far from the tipping, the regression based on higher order polynomials leads to wrong results. This strengthens the argument for the fail-mechanism already described, i.e., that higher order terms get noise-based gradients assigned to them as they are not varying significantly during training and regression. When they then do start to vary during tipping, they will give large yet wrong contributions to the weight prediction and this way will deteriorate the child model and its output.

In general, the skill of all models (parents and children) improves with larger EOT. This is not only due to more training data per se, but rather that the training data then also includes information located closer to the tipping point. Consequently, the models that are trained up-until  tipping perform best, i.e., all yellow boxes go down from left to right in each panel of Fig.~\ref{fig:EOT_split}. On average, even as children and parents show increasingly similar performances with EOT closer to the tipping point, on average the better RMSE of the child models are never reached by the parent models.

A note on temporal data: In experiment 1, we shuffle the training data before applying the training/validation split. In some problems this is not possible, either due to the choice of architecture (as LSTM) or not advised due to correlations within time \citep[e.g.,][]{Schnaubelt.2019,Bergmeir.2018}. If data shuffling is not an option but a test/validation split is still necessary, then the training data is pushed even further away from the tipping point by the presence of the validation window. As a result, the parent models will show an even worse performance. Consequently, our weight-prediction approach will become more advantageous as it also uses the validation data located closer to the tipping point for the weight-regression.

\subsection{Experiment 2: Estimation of Sea Water Density}
The NN of choice here is a simple two layer feed-forward NN with ELU activation functions in the first layer and linear activation in the final layer. The regression of the NN weights uses the same set of variables as the NN input, i.e., S, T, P, $\phi$, $\lambda$, and z (cf Sec.~\ref{sec:exp2}). For the forgetful online learning (Tab.~1, step 2a), we finetune the parent model only on individual $x_i$ without the regularization towards the entire training data set $\mathbb{I}$.

In experiment 2, the training data set is very large and thus retraining the parent-model on every training data point is prohibitively expensive. A random sub-sampling of the training set could lead to undersampling with respect to some of the predictors if the random draw is not sufficiently large. To be efficient and at the same time avoid undersampling the predictors, we sort the training data into quantiles (e.g., 300) for every predictor. Then we draw several $x_i$ from every parameter-specific quantile-interval for finetuning the parent model (Tab.~1, step 2). If it is suspected that special functions (or interactions) of predictors should play a role in the weight-regression, then sampling quantiles of the respective function-values (or interactions) is advised (e.g., one has not only to guarantee sufficient sampling of the predictors x and y, but also of log(x), y$^2$, and x*y etc.). 

As in experiment 1, individual \textit{ChildModel$_t$} are generated for every single target data point $x_t$, here every grid point of the 3d ocean grid.

\begin{figure}
\includegraphics[scale=0.3,trim=0cm 0cm 1.5cm 0.96cm,clip=true]{
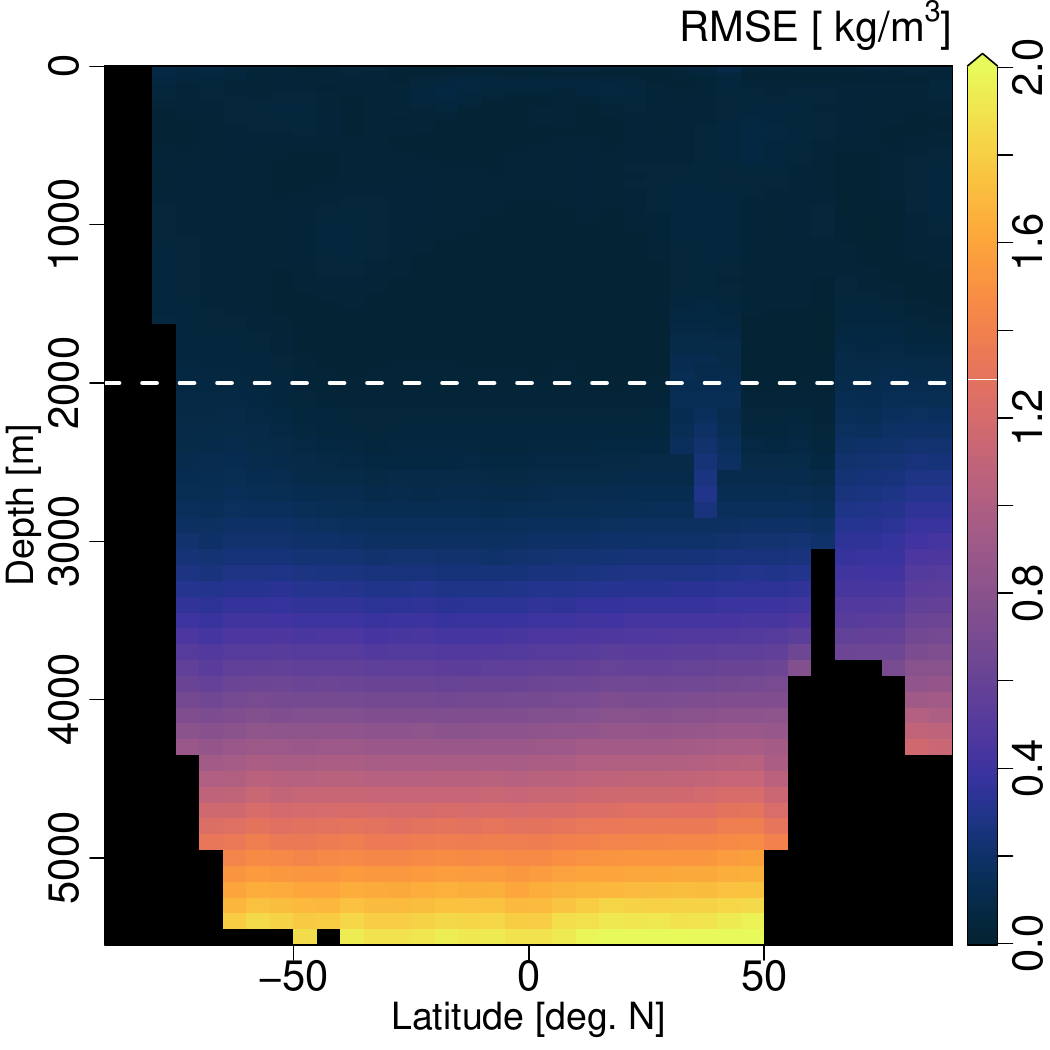}%
\includegraphics[scale=0.3,trim=1.6cm 0cm 0cm 0cm,clip=true]{
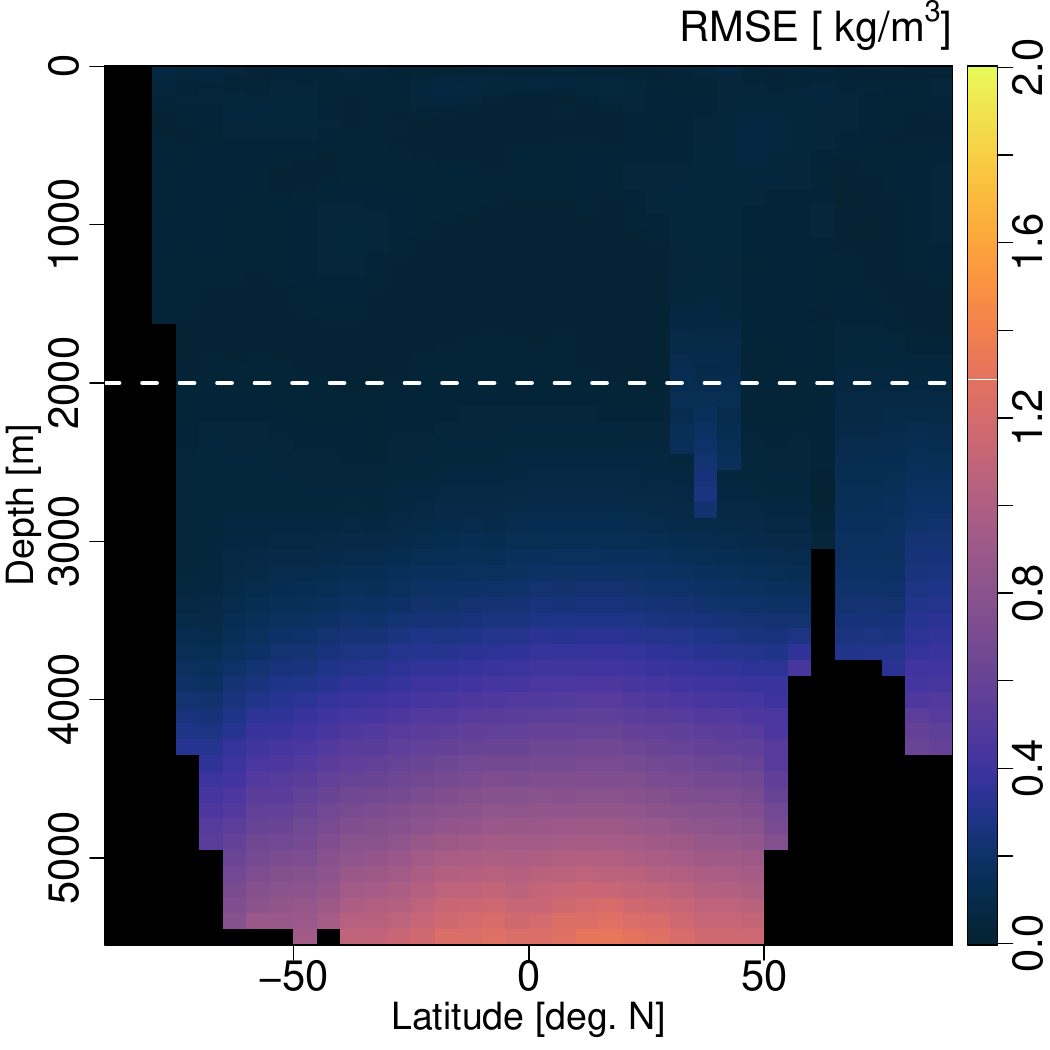}
\includegraphics[scale=0.3,trim=1.6cm 0cm 0cm 0cm,clip=true]{
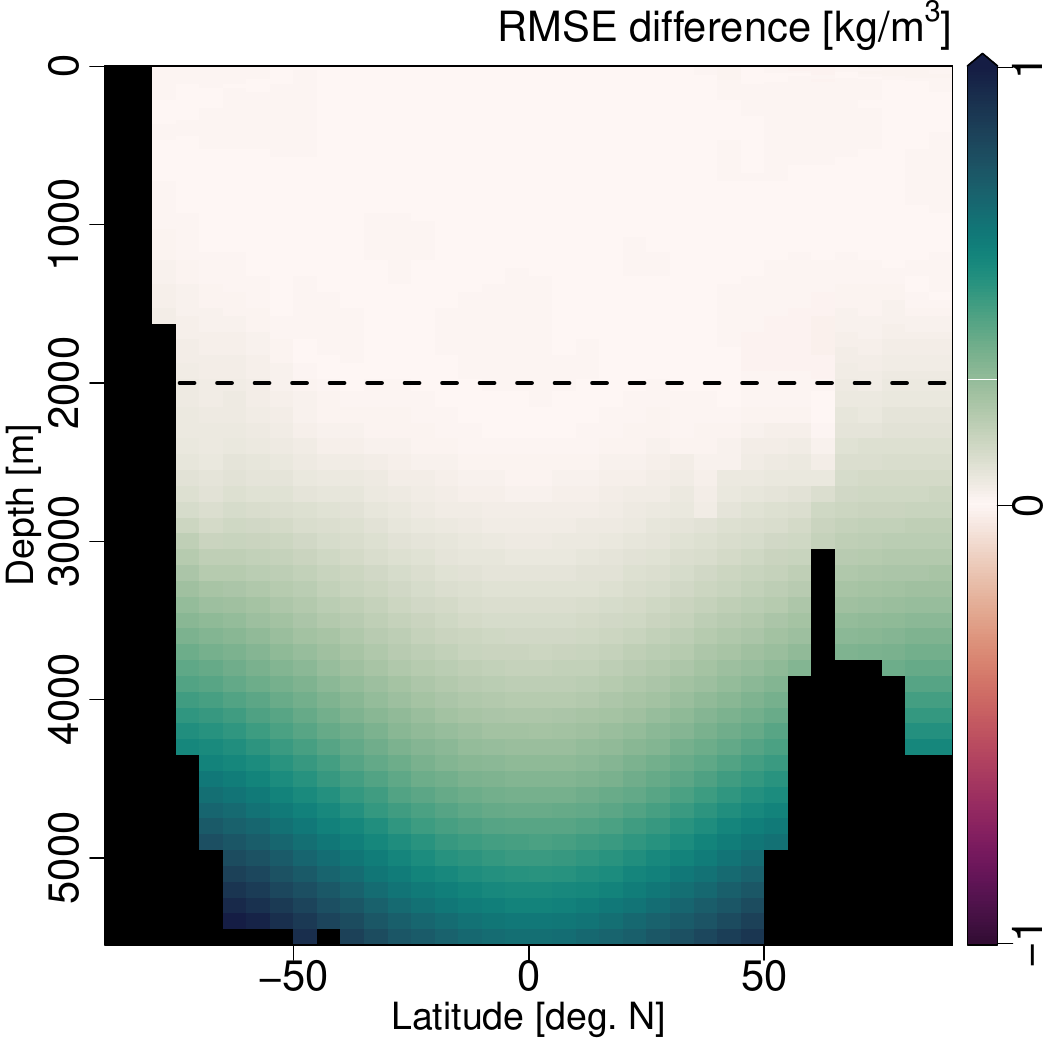}\\%
\includegraphics[scale=0.3,trim=0cm 0cm 1.5cm 0.96cm,clip=true]{
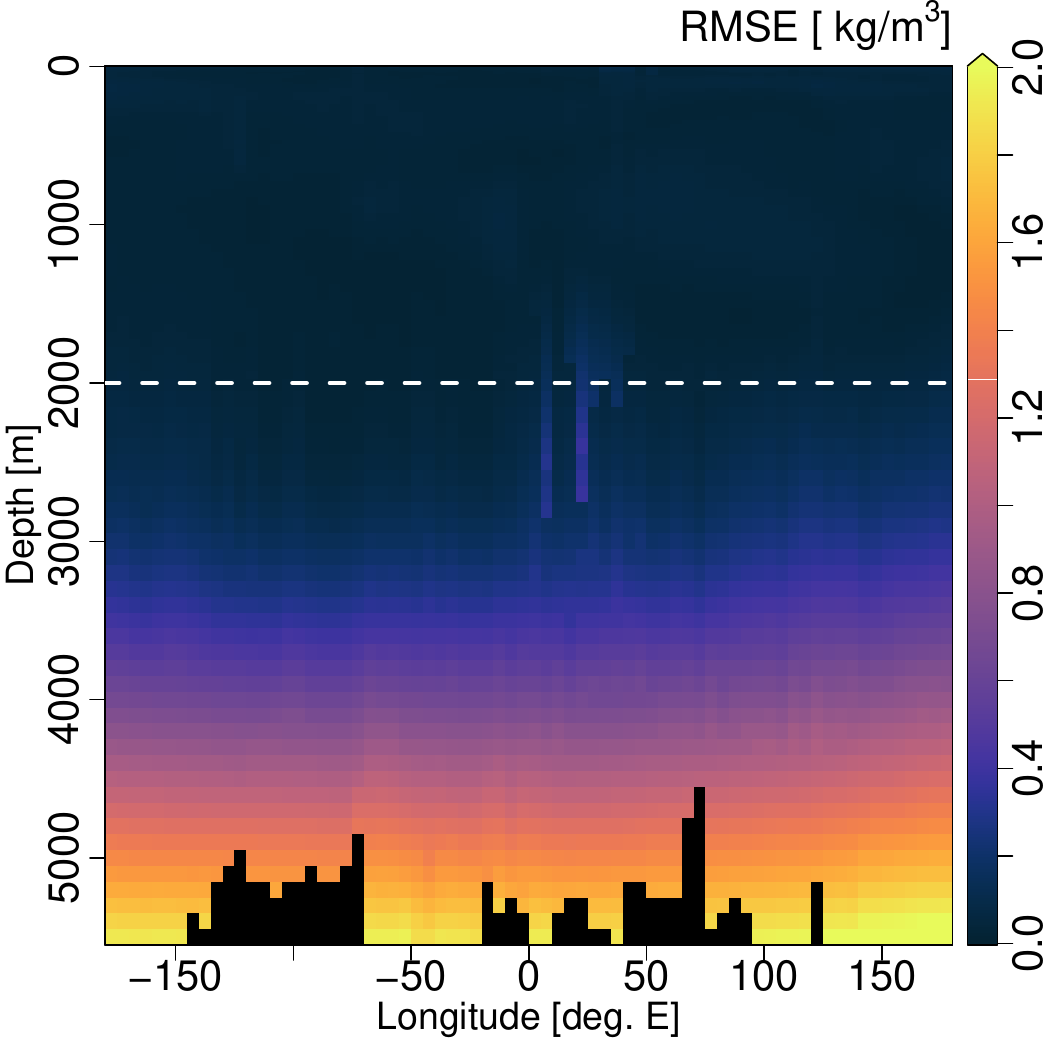}%
\includegraphics[scale=0.3,trim=1.6cm 0cm 0cm 0cm,clip=true]{
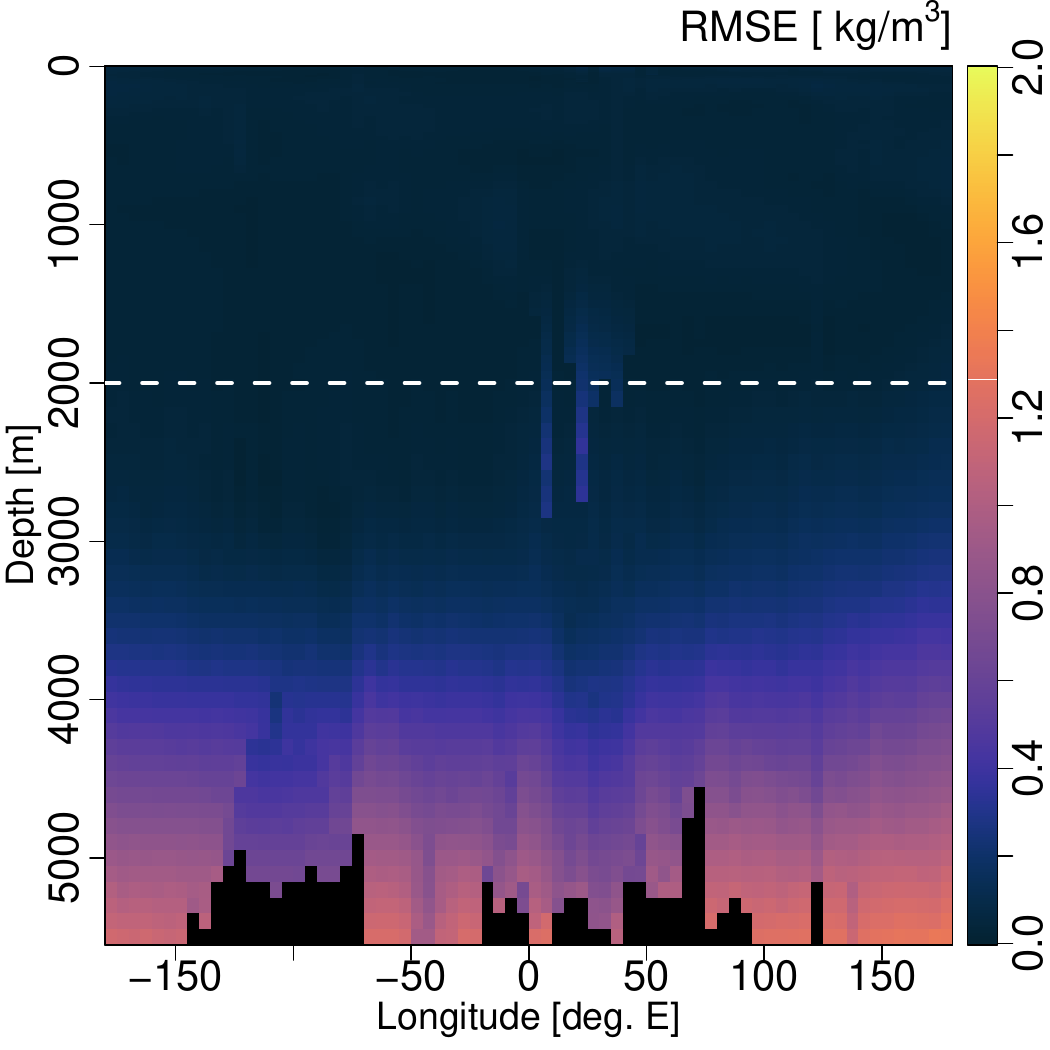}
\includegraphics[scale=0.3,trim=1.6cm 0cm 0cm 0cm,clip=true]{
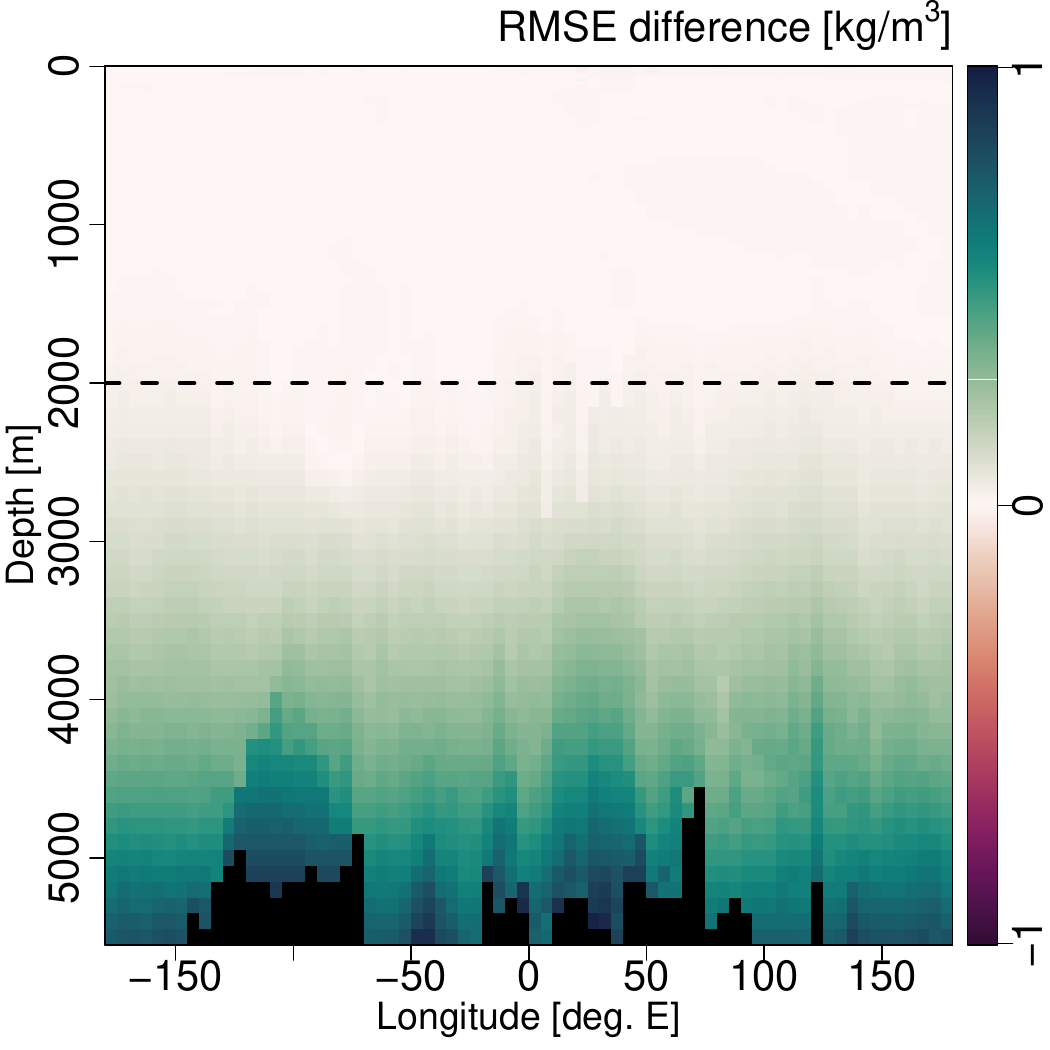}%
\caption{Depth dependent zonally averaged$^*$ (upper panels) and meridionally averaged$^*$ (lower panels) RMSE of NN based sea water density estimation after training above the dashed line only. Left panels: Performance of the parent model. Middle panels: Performance of weight-predicted child model. Right panels: RMSE differences (parent minus child, the green color represents improvements by the child model). $^*$: In giving every global output location equal weight, these averages represent a data-centered view and do not follow oceanographic standards.}
    \label{fig:EOS_depth_poly1} 
\end{figure}

Figure~\ref{fig:EOS_depth_poly1} shows the impact of the weight-prediction approach on a spatial OOD problem, where the parent model is only trained above the dashed line at 2000 m.

As a result, the weight-prediction approach improves the performance of the NN sea water density calculation in the deep ocean considerably. At depth, the improvements amount to 2 kg/m$^3$ in the global average, which is substantial by oceanographic standards. Below the EOT of 2000 m, the child model improves the RMSE of the parent model by 50-100\%.  Regions where the parent model already performs well show less or no improvements. As already discussed in experiment 1, even slightly worse RMSE can be found if the approach is applied to the training data. In contrast to experiment 1, the result of experiment 2 appear to be very stable (not shown).

\subsection{Experiment 3: Uncertainty Estimation of Global Wind Velocity Reanalyses}
As outlined in Sec.~\ref{sec:exp3}, this experiment assesses the proposed weight-prediction framework in the context of global wind velocity uncertainty estimation. The focus is on evaluating whether knowledge about uncertainty characteristics learned at one oceanic location can be extrapolated to terrestrial regions with different atmospheric dynamics. This task represents a demanding OOD test case for the proposed method.

Unlike the previous experiments, the child model generation (cf. Tab.~1, steps 3 and 4) is conducted by a NN, too. This choice of a NN-based child model generation underscores the significance of the underlying weight-prediction methodology over regression model architectures. The child model generation is implemented as a NN consisting of two fully connected hidden layers with 32 nodes each and parameter-specific output heads. Instead of predicting all network weights and biases of the child model, we restrict the weight-regression and prediction to the final fully connected layer, while keeping the remaining parameters identical to those of the parent model. For this to work correctly, the not-to-be-predicted weights and biases have to be set as \texttt{trainable=FALSE} already during the online-learning step (Tab.~1, step 2). The use of multiple output heads, together with a scaled mean-squared-error (MSE) loss, mitigates scaling disparities across the predicted weights and biases. ELU activation functions and the Adam optimizer (learning rate = 0.001) are used. 

The training and evaluation periods span 2012–2017, using six-hourly data from two reanalysis products (ERA5 and CFSv2, cf. Sec.~\ref{sec:exp3}). Data from 2012–2016 are used for training, January–June 2017 for validation, and July–December 2017 for testing. All results reported below correspond to this testing period. From each of the two input data sets, the principal components of the four leading EOF, yielding eight predictor values in total per time step, are used as weight-regressors. For the forgetful online learning (Tab.~1, step 2a), n is set to 200. The parent model is trained at a single oceanic grid point located at 180° E, 60° S, while the child model predictions are evaluated over the entire land mass to assess the method’s ability to generalize beyond the training domain. As in the other experiments, individual \textit{ChildModel$_t$} are generated for every single target data point $x_t$, that is every grid point of the 2d globe.

\begin{figure}
    \centering
    \includegraphics[width=\textwidth]{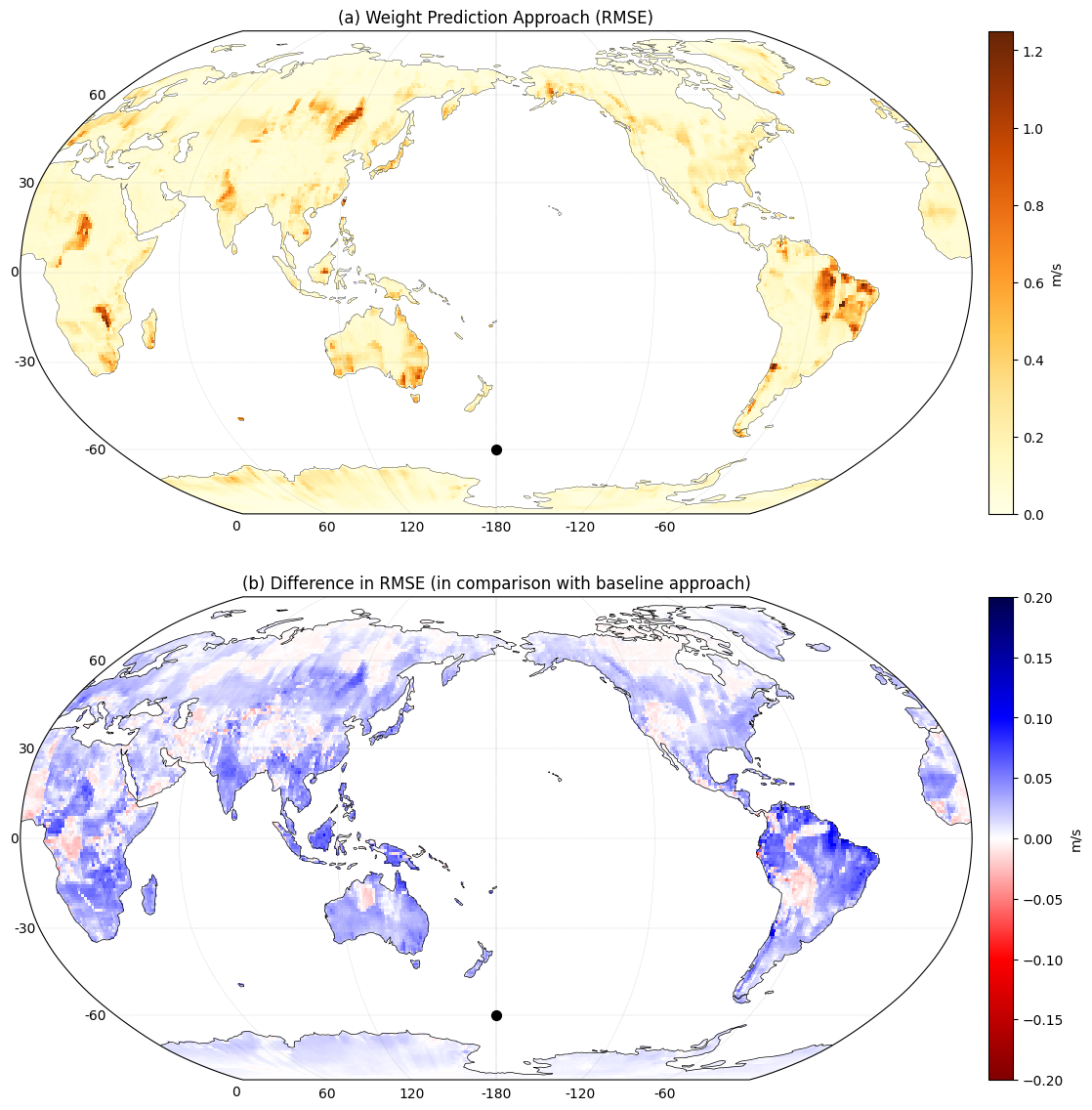}%
    \caption{Spatial distribution of RMSE in the prediction of wind-speed uncertainty using the proposed weight-prediction framework. (a) RMSE of the weight-predicted neural network trained at the reference oceanic location (180° E, 60° S), represented by black dot. (b) Difference in RMSE between the parent model and the proposed method, where blue regions indicate improvement.}
    \label{fig:wsu_nn_batched}
\end{figure}

Figure~\ref{fig:wsu_nn_batched} summarizes the spatial distribution of the RMSE in wind speed uncertainty prediction. Panel (a) shows the RMSE obtained with the weight-predicted model trained only at the oceanic reference point (black dot) and panel (b) illustrates the parent minus child model RMSE differences. Blue areas in Fig.~\ref{fig:wsu_nn_batched}(b) mark regions where the proposed method outperforms the parent model, i.e., where RMSE is reduced by the weight-prediction method.

From these results, it is evident that the weight-prediction approach substantially enhances the wind velocity uncertainty estimation over continental regions, particularly where the parent model struggles the most. Notable improvements are observed over the southeast coast of Australia, eastern Brazil, the East African Highlands, the Indonesian archipelago, and southeastern Russia. In contrast, regions where the parent model already performs well show no major improvement, as expected. This applies for most of the oceanic region as well (not shown in the figure). This behavior, suggests that the method provides the greatest benefit under conditions of strong distributional shift, where the parent model’s learned representations are least transferable (cf. experiment 1 and 2).

Quantitatively, the global mean RMSE of the uncertainty estimates decreases from 0.1236 m/s to 0.1104 m/s, corresponding to an average improvement of approximately 10.7\%. Repeated experiments indicate that this value varies between -2\% and 14\%, with a mean improvement of 5.25\%, reflecting again the stochastic variability inherent in NN training. Despite this variability, the consistent tendency across repeated experiments confirms that the proposed method systematically improves the predictive performance of the parent model, particularly in regions where generalization is most challenging.

That a NN can be used to improve the OOD problem of another NN seems contradictory at best. One could ask why the child-generating NN is not limited by OOD, and why then not one NN alone can handle the problem. A full answer to these questions remains open to further research. So far, we only can argue that we as user know that in experiment 3 the task we gave to the NN remains generally the same but has to be adapted outside the training realm. Consequently, splitting an OOD-experiment into problem-solving and problem-adaptation represents a form of structural implementation of prior knowledge. This understanding, the required task separation, and the subsequent training of both NN, a single NN cannot (yet) develop itself.

\section{Summary and Final Remarks}
In this study, a general method is presented that aims at improving neural network (NN) performance outside of their training data distribution (OOD). The method is proposed for problems where re-normalization or standardization is not possible due to limited data in few-shot problems or due to governing nonlinear relations, e.g., in climate sciences.
Additionally, OOD is meant in a wider sense here. Even when all input variables have the same distribution within and out of training, a trained NN can operate on uncharted territory due to unseen value-combinations and interactions of the input variables.

After the training of a NN, which we call "parent model",  on a specific problem, the method requires the following three steps: 
First, the method collects the series of weight anomalies that arise by  online-relearning or fine-tuning within the training data. Second, a relation (e.g., a linear regression) is to be established between these weight anomalies and suitable predictors (e.g., the input variables of the training data set itself). Third, the established relation is then used to generate new weights and thus a new NN that correspond to an application data set by employing the same predictors but now with values that correspond to the application data set (e.g., the application data itself). In analogue to the parent model, we call this new model the "child model". A child model can encompass a group of NN as a new NN can be generated for every single application data point or location. However, sub-setting is possible in all steps of the approach. For example, only one new NN could be generated per cluster of the application data set. Likewise, the weight sensitivities could be derived from quantiles or clusters of training data only. 

In short, the approach is generating NN fine-tuned on subsets of the training data and then is extrapolating along these NN to data-regions outside of the training data  distribution. This way, network's learned tasks can be transferred, e.g., to different geographic locations, future climates, or across (physical) regimes. 

The method is demonstrated on 3 very different examples from Earth sciences, one temporal problem, one spatial problem, and one spatiotemporal problem. These examples and their execution demonstrate further the possible flexibility in the approach in using linear, nonlinear, and NN-based regression methods; full and partial NN weight prediction; and different realizations of the online training step.
Despite the fact that all given examples are from Earth sciences, the authors strongly believe that the demonstrated approach is useful in other fields, as astrophysics, biology, social sciences, or technical applications, too. For example, in medicine it could be used to transfer learned skills and knowledge to target groups underrepresented or unrepresented in the training data.  

In summary, our experiments demonstrate that the proposed weight-prediction framework enhances the extrapolation capability of NN under strong temporal, spatial, and domain shifts. The method proves most effective where traditionally trained and therefore static models exhibit limited skill, underscoring its potential to extend NN capabilities to previously unseen regimes and conditions.
As a result, the NN become adaptive and dynamical operators.

Despite all these promising results, there are still some caveats, as choices have to be made that can heavily impact the method's performance. The choice of suitable predictors, e.g., if one does not choose the input variables themselves, have to be based on prior knowledge. This should pose no major problem for the experienced researcher. The same can be stated about the choice of whether only specific parts of a NN or all weights and biases should be subject to the weight-regression and subsequent weight-prediction.
In contrast to these rather straightforward choices, the choice of the weight-regression and prediction method cannot be based on prior domain knowledge alone. The reason is, that the weight sensitivities the proposed approach has to model are entwined with the network architecture and the employed activation functions as well. As a first guide, linear or slightly nonlinear regression methods should work fine for most cases.

The biggest drawback identified so far is that the weight-prediction method has inherited the notorious stochastic nature of NN training. From several parent NN (with identical architecture and hyper-parameters) that are trained on the same data, not all do produce better performing offspring. To this issue, the most follow-up research will be dedicated. 
Although we already discussed possible reasons and solutions within the manuscript, we will have to understand the problem better to be able to identify and select parent models suitable for weight-regression or in the long run even foster suitable parent model generation through tailored training procedures. Additionally, we see room to improve the generation of the child models by predictor pruning, weight-prediction validation, regression-ensembles, transformers, or other means.  

As a conclusion, most steps of the approach may be called more or less experimental and surely can and will be improved in follow-up research. However, already now the method has to be considered very promising and is at least in our experiments robust in the ensemble-average sense.

As a further outlook, we see this method also as a promising tool for NN applications without OOD. During our many tests, we got the impression that the method can improve the performance of undertrained, overfitted, or by design under-complex models as well.

\section*{Acknowledgments}
JSW receives funding from the German Federal Ministry of Education and Research (BMFTR) through the project PalMod. 
SRS receives funding from the Helmholtz International Berlin Research School in Data Science (HEIBRiDS) and the Helmholtz Foundation.

\section*{Authors contribution statement}
JSW contributed with  methodology development, experiment design, implementation, and interpretation. JSW wrote the first draft of the manuscript, did finance the project, and did project 
supervision. SRS contributed with experiment design, method development, implementation, and refining the manuscript.

\section*{Competing interests}
The authors declare no competing interests.

\bibliographystyle{apalike}

\end{document}